%% file: acl_latex.tex
\documentclass[11pt]{article}

\usepackage[final]{acl}
\usepackage{amsmath} %自己加的
\usepackage{amsfonts} %自己加的
\usepackage{algorithm}%自己加的
\usepackage{algorithmic}%自己加的
\usepackage{booktabs}%自己加的
\usepackage{multirow}%自己加的
\usepackage{times}
\usepackage{latexsym}

\usepackage[T1]{fontenc}
\usepackage[utf8]{inputenc}

\usepackage{microtype}

\usepackage{inconsolata}

\usepackage{graphicx}

\title{SpecBound: Adaptive Bounded Self-Speculation with Layer-wise Confidence Calibration}

\author{
    Zhuofan Wen\textsuperscript{\rm 1,2,3},
    Yang Feng\textsuperscript{\rm 1,2,3}\footnotemark[2] \\
    \textsuperscript{\rm 1}{Key Laboratory of Intelligent Information Processing, Institute of Computing Technology,} \\ Chinese Academy of Sciences (ICT/CAS) \textsuperscript{\rm 2} {State Key Laboratory of AI Safety,} \\ Institute of Computing Technology, Chinese Academy of Sciences \\
    \textsuperscript{\rm 3} {University of Chinese Academy of Sciences, Beijing, China} \\
    \texttt{\href{mailto:wenzhuofan24z@ict.ac.cn}{wenzhuofan24z@ict.ac.cn},  \href{mailto:fengyang@ict.ac.cn}{fengyang@ict.ac.cn}}
}
\begin{document}
\maketitle
\renewcommand{\thefootnote}{\fnsymbol{footnote}} %将脚注符号设置为fnsymbol类型，即特殊符号表示
\footnotetext[2]{Corresponding author: Yang Feng.} %对应脚注[1]

\renewcommand{\thefootnote}{\arabic{footnote}}
\begin{abstract}
Speculative decoding has emerged as a promising approach to accelerate autoregressive inference in large language models (LLMs). Self-draft methods, which leverage the base LLM itself for speculation, avoid the overhead of auxiliary draft models but face limitations: shallow layers often produce overconfident yet incorrect token predictions, and the presence of difficult tokens in a draft sequence forces redundant computation through deeper layers, undermining both draft acceptance and overall speedup. To address these issues, we propose a novel self-draft framework that suppresses spurious confidence via layer-wise temperature annealing in early-exit decision and adaptively bounds speculation length based on token-wise decoding difficulty. By reprocessing the hidden states of draft tokens in a unified parallel pass through deep layers, our method maintains exact output equivalence with the original model while maximizing computational efficiency. It requires no modifications to the base LLM parameters and achieves up to \textbf{2.33×} wall-time speedup over standard autoregressive decoding across diverse long-form generation tasks and multiple model architectures.\footnote{{\url{https://github.com/ictnlp/SpecBound}}}
\end{abstract}

\input{tex/1-introduction}
\input{tex/3-method}
\input{tex/4-experiment}

\input{tex/2-background}
\input{tex/5-conclusion}
\input{tex/6-limitation}
\input{tex/8-Acknowledgements}


\bibliography{custom}
\begin{figure*}[t]
  \includegraphics[width=1\linewidth]{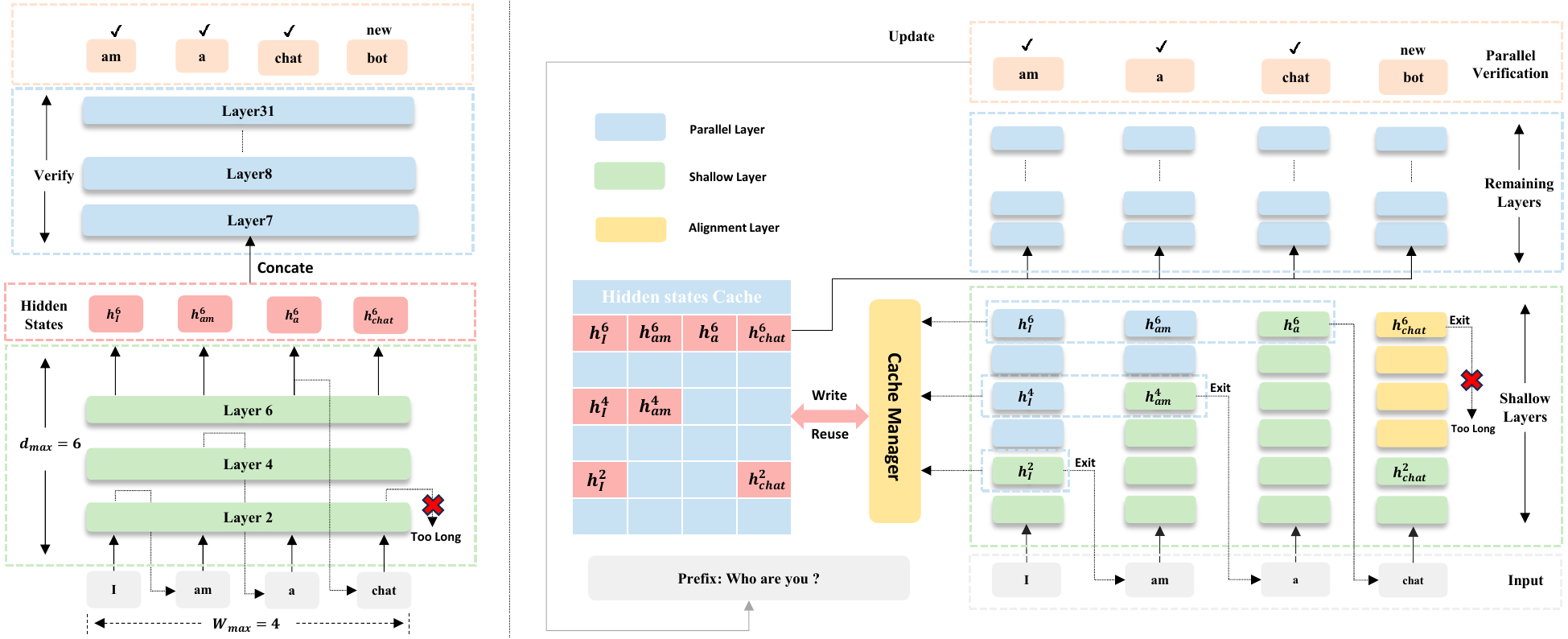}
  \caption{\textbf{Left:} Overview of the BSCS algorithm under the setting $d_{\max}=6$ and $w_{\max}=4$. When token ``chat'' successfully exits at layer 2 and the total draft length reaches the width bound $w_{\max}$, speculation is terminated. All hidden states of previously generated tokens (``I'', ``am'', ``a'', ``chat'') are concatenated and passed to the remaining layers for parallel verification. \textbf{Right:} Detailed illustration of per-token layer-wise computation and hidden-state cache management.}
    \label{wmax architecture}
\end{figure*}
\newpage
\appendix

\input{tex/7-appendix}

% \label{sec:appendix}

% This is an appendix.

\end{document}

%% file: tex/1-introduction.tex
\section{Introduction}

Large Language Models (LLMs) excel at diverse text generation tasks \cite{achiam2023gpt,guo2025deepseek}, yet practical applications increasingly require long outputs, such as chain-of-thought reasoning \cite{wei2022chain}, multilingual capability \cite{bu2026language} and tool coordination in AI agents \cite{shen2023hugginggpt}. For standard autoregressive decoding, the combination of these long sequences and ever-larger model sizes leads to high inference latency, motivating the need for more efficient parallel decoding strategies \cite{geng2021learning,nie2025large,zhang2025survey}. Among these, speculative decoding (SD) has been proposed as a practical framework for accelerating inference \cite{leviathan2023fast,chen2023accelerating}. 

Early SD methods rely on an independent draft model—a smaller, separately trained network that proposes candidate tokens—requiring additional model selection and training overhead \cite{cai2023medusa,li2024eagle,xia2023speculative,wen2024speculative}. To eliminate this dependency, recent work has shifted toward self-draft strategies that leverage the base LLM itself for speculation. These approaches broadly fall into three categories: (1) fine-tuning, where the LLM is adapted with non-autoregressive objectives to enable parallel token prediction, but at the risk of degrading generalization due to parameter updates \cite{lin2025bita,yi2024generation,gloeckle2024better}; (2) layer skip, which accelerates draft generation by computing only a subset of layers, yet often compromises output quality and disrupts KV cache consistency \cite{zhang2024draft}; and (3) early exit, which generates draft tokens from shallow layers of the base LLM and feeds them into subsequent decoding steps. While early exit ensures lossless output, current implementations achieve limited speedup \cite{elhoushi2024layerskip,liu2024kangaroo}. \begin{figure*}[t]
  \includegraphics[width=1\linewidth]{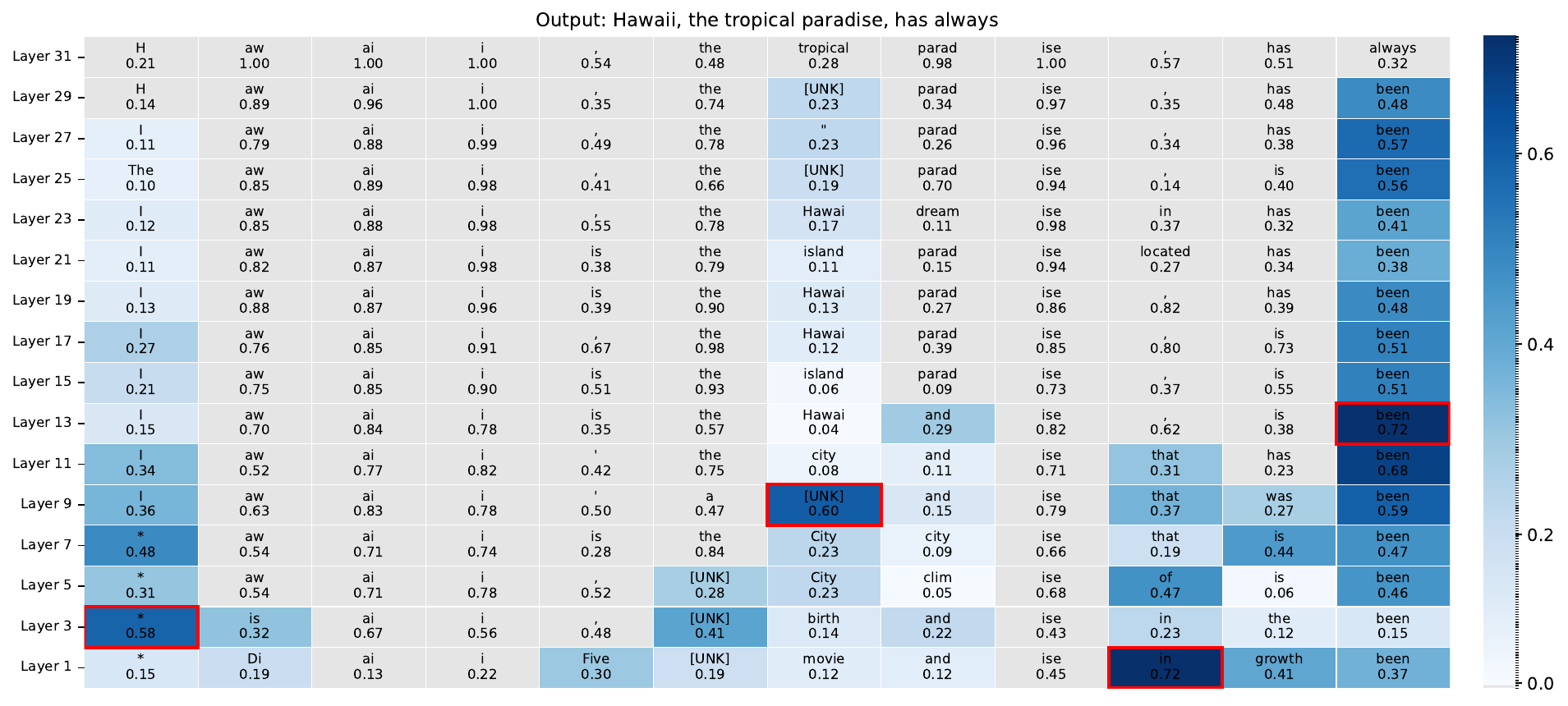}
    \caption{
    Layer-wise token prediction and confidence of a segment with the MT-Bench prompt. Each cell shows the greedy-decoded token and its confidence (color intensity: darker = higher probability) at a given layer. Gray cells indicate that the token has already correctly exited at an earlier layer and is no longer considered for early-exit decisions. This figure reveals the issues of spurious high-confidence predictions and heterogeneous exit depths.
    }
    \label{fig:layer-wise analysis}
\end{figure*}

Through visualization of intermediate layer dynamics in Figure \ref{fig:layer-wise analysis}, we observe significant heterogeneity in token-level decoding difficulty \cite{xia2025tokenskip}. While most tokens can correctly exit at shallow layers, a small fraction of difficult tokens require deeper computation to resolve semantic ambiguity. Due to the batched verification requirement in speculative decoding, these difficult tokens force the entire draft sequence to be processed through deeper layers, leading to redundant computation. Moreover, pretraining loss function supervises only the final-layer output, leaving shallow layers without direct optimization signals. As a result, shallow layers often assign spurious high confidence to incorrect tokens—a tendency that easily misleads naive early exit methods relying solely on confidence thresholds, severely reducing draft acceptance and limiting acceleration.

To address the above challenges, we propose SpecBound, a self-draft speculative decoding framework featuring two key components. First, to suppress spurious high-confidence predictions in shallow layers, we introduce the Annealed Confidence Threshold early exit mechanism, which applies a layer-dependent temperature schedule during early-exit decisions to smooth softmax outputs and attenuate false peaks. Second, to mitigate redundant computation caused by difficult tokens, we design the Bounded Speculation with Cached States algorithm: during autoregressive draft generation, if any token fails to exit before reaching the maximum depth or the maximum draft length, the current speculation phase is terminated; cached hidden states of all draft tokens are then jointly processed through the remaining layers in a single parallel pass, ensuring lossless output while maximizing parallel efficiency.

Our contributions are as follows:
(1) We propose Annealed Confidence Threshold, a more reliable early-exit criterion that suppresses spurious confidence in shallow layers via layer-wise temperature annealing;
(2) We design Bounded Speculation with Cached States algorithm that coordinates speculation and verification through adaptive depth and width control;
(3) SpecBound achieves up to \textbf{2.33$\times$} wall-time speedup over standard autoregressive decoding across diverse long-form generation tasks and multiple base models, requiring no modification to the base LLM parameters and producing outputs identical to the original model.

%% file: tex/3-method.tex
\section{Methods}
In this section, we present our proposed acceleration framework, SpecBound. We begin by visualizing the intermediate layer computation of LLM in Section~\ref{subsec:Layer-wise Decoding Analysis}, which reveals two key patterns observed in our preliminary experiments. Guided by these insights, we introduce the temperature-annealed early-exit mechanism in Section~\ref{subsec:Annealed Confidence Threshold} and the Bounded Speculation with Cached States Algorithm in Section~\ref{subsec:Bounded Speculation with Cached States}. Finally, we provide a theoretical wall-time speedup analysis to demonstrate the effectiveness of our method in Section \ref{subsec:Wall-time Speedup Analysis}.
\subsection{Layer-wise Decoding Analysis}
\label{subsec:Layer-wise Decoding Analysis}
\paragraph{Layer-wise Visualization}
While several recent works have explored early-exit-based approaches for lossless acceleration, their speedup remains limited and currently falls short of the performance achieved by mainstream independent-draft methods. To investigate the factors that constrain the acceleration potential of early exit, we analyze a representative example from MT-Bench which requires generating a blog on traveling in Hawaii.

For each token during generation, we probe the output of every Transformer layer in the base LLM by projecting its hidden state through the language model (LM) head and applying greedy decoding to obtain the most likely token at that layer \cite{chuang2023dola,zhu2025layercake}. Notably, to account for the poor predictive capability of shallow layers, we follow AdaDecode~\cite{wei2025adadecode} and train lightweight intermediate-layer LM heads for early-exit decisions. Figure~\ref{fig:layer-wise analysis} visualizes the predicted token and its confidence at each layer, from which we identify two distinct early-exit patterns that appear to influence the achievable speedup.

\paragraph{Spurious High Confidence in Shallow Layers} First, for those challenging tokens that ultimately require deep layers to be correctly predicted (e.g., token ``H'' and ``tropical''), shallow layers often exhibit spuriously high confidence in incorrect tokens which is marked by red boxes in figure \ref{fig:layer-wise analysis}. We hypothesize that this stems from the pretraining objective of base models: since the loss function supervises only the final-layer output, shallow layers lack direct optimization signals and thus may develop overconfident but incorrect predictions. Since many existing early-exit methods rely on simple threshold-based criteria, these overconfident yet erroneous predictions can easily pass the exit condition, leading to low-quality drafts.

\paragraph{Heterogeneous Token Difficulty} Second, as shown in Figure~\ref{fig:layer-wise analysis}, decoding sequences exhibit significant heterogeneity in token-level difficulty. While a consecutive span of tokens, from ``aw'' to ``the'', stabilizes correctly at shallow layers, a few semantically challenging tokens, such as ``H'', ``tropical'', and ``always'', remain uncertain through shallow-to-mid layers and require deep computation. Due to the batched verification requirement in speculative decoding, the presence of even one such difficult token forces the entire draft sequence to be processed through deeper layers. This leads to two sources of inefficiency: first, when the draft is accepted, simple tokens incur redundant deep computation that erodes their early-exit speedup potential; second, when rejected, the substantial time spent computing deep representations for the difficult token becomes wasted overhead, directly slowing down end-to-end decoding.

% These observations highlight two fundamental limitations of current early-exit strategies: poor draft quality due to overconfident shallow-layer predictions and inefficient speculation caused by heterogeneous exit depths across tokens. To address these issues, SpecBound introduces two complementary mechanisms, which is detailed in the following subsections.

\subsection{Annealed Confidence Threshold}
\label{subsec:Annealed Confidence Threshold}
To address the issue of spurious high-confidence predictions in shallow layers, we propose the Annealed Confidence Threshold(ACT) early exit mechanism. Unlike prior early-exit methods that determine exit solely by comparing the probability of the greedy-decoded token against a fixed threshold, ACT dynamically modulates the sampling temperature based on layer depth and couples it with a fixed confidence threshold for early-exit decisions, suppressing premature exits caused by false certainty.

Specifically, considering a base LLM that contains $L$ transformer layer, we scale the early exit logits $\mathbf{z}^{(\ell)}$ at layer $\ell$ ($1 \leq \ell \leq L$) with a layer-dependent temperature $T_\ell$:
\begin{equation}
  \label{eq:ACT1}
  T_\ell = 1 + \alpha \left(1 - \frac{\ell}{L}\right)
\end{equation}
\begin{figure*}[t]
  \includegraphics[width=1\linewidth]{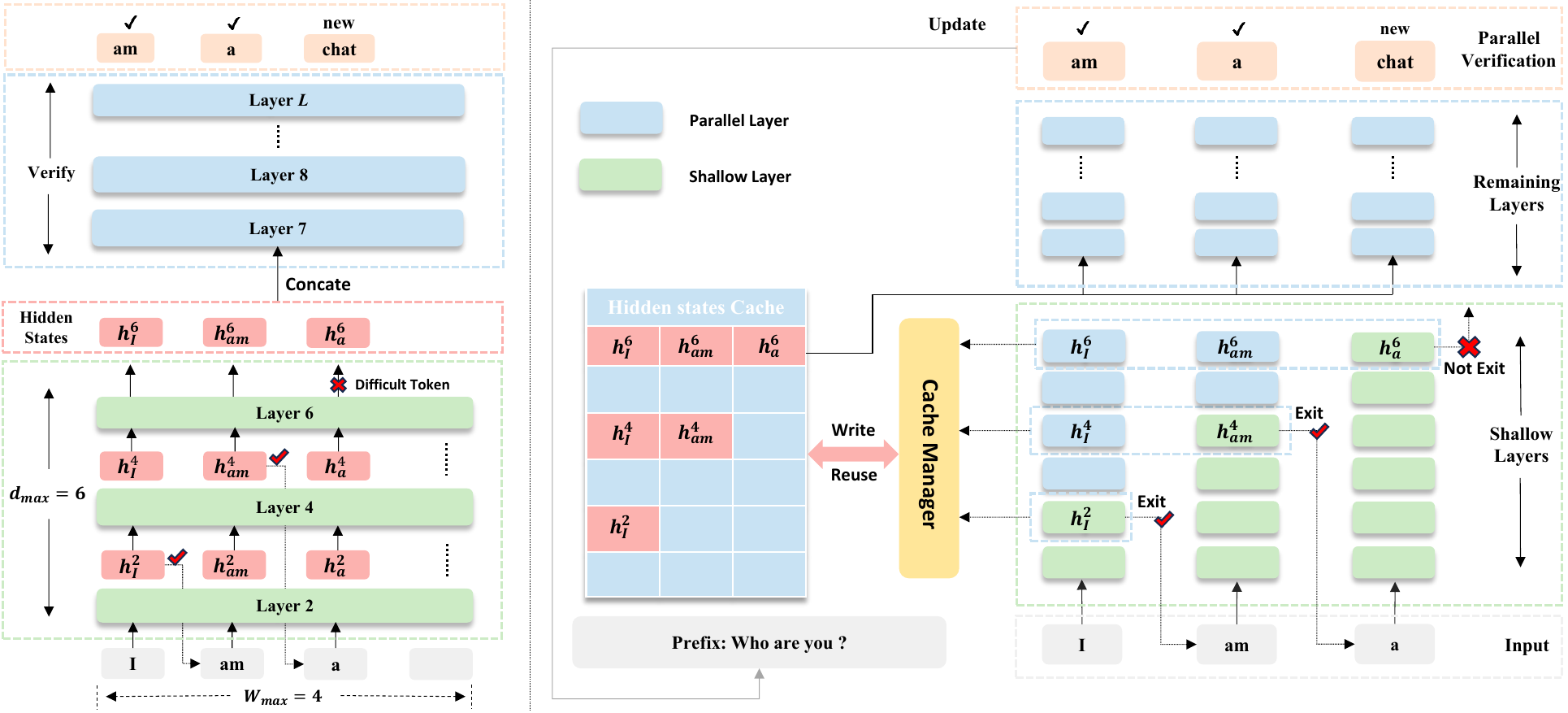}
  \caption{\textbf{Left:} Overview of the BSCS algorithm under the setting $d_{\max}=6$ and $w_{\max}=4$. When token ``a'' fails to exit by layer 6, speculation is terminated. All hidden states of previously generated tokens (``I'', ``am'', ``a'') are concatenated and passed to the remaining layers for parallel verification. \textbf{Right:} Detailed illustration of per-token layer-wise computation and hidden-state cache management.}
  % \caption {A minimal working example to demonstrate how to place
  %   two images side-by-side.}
  \label{dmax architecture}
\end{figure*}
where $L$ is the total number of layers in the base LLM and $\alpha > 0$ controls the strength of annealing. This linear schedule ensures higher temperatures in shallow layers (e.g., $T_1 = 1 + \alpha$) to flatten the output softmax distribution and reduce overconfidence, while gradually cooling to the standard temperature $T_L = 1$ at the final layer which ensuring that the output distribution remains identical to that of the original base LLM. With adaptive sampling temperature, the confidence of the top-1 token at layer $\ell$ is then computed as:
\begin{equation}
  \label{eq:ACT2}
    p^{(\ell)} = \max\left( \mathrm{softmax}\left( \frac{\mathbf{z}^{(\ell)}}{T_\ell} \right) \right)
\end{equation}

Early exit is triggered at the first layer $\ell$ where $p^{(\ell)} \geq \tau$, with $\tau \in (0,1)$ being a predefined threshold. By raising the temperature in shallow layers, ACT effectively lowers the raw confidence $p^{(\ell)}$ for incorrect tokens that would otherwise appear overconfident under $T=1$, making it harder for them to satisfy the exit condition. This results in more reliable draft tokens without altering the base model’s parameters and incurs only negligible computational overhead, as temperature scaling is a lightweight post-activation operation.

\subsection{Bounded Speculation with Cached States}
\label{subsec:Bounded Speculation with Cached States}
To address the issue of heterogeneous exit depths observed in prior work, we introduce the Bounded Speculation with Cached States Algorithm (BSCS) Algorithm, which explicitly limits both the depth and width of speculative generation. We define two hyperparameters:
 $d_{\max}$: the maximum layer depth allowed for early-exit draft generation;
 $w_{\max}$: the maximum number of consecutive tokens that can be drafted before triggering verification.This dual-bound design ensures predictable speculation latency and prevents excessive computation on difficult tokens or error-prone long draft sequences.Figure~\ref{dmax architecture} illustrates a complete cycle of BSCS’s draft-verify process using a concrete example with $d_{\max}=6$ and $w_{\max}=4$ for clarity, while the actual parameter settings used in our evaluation are reported in Section~\ref{Experimental Setup}.

\paragraph{Handling Difficult Tokens}
Given a decoding history prefix ``Who are you?'', the base model early exits token ``am'' at layer 2 and token ``a'' at layer 4 at the first two steps. In the third step, the model attempts to draft the next token but fails to meet the early-exit condition mentioned in Section~\ref{subsec:Annealed Confidence Threshold} even after computing up to the depth bound $d_{\max}=6$, signaling a difficult token that requires deeper processing. At this point, BSCS immediately terminates the current speculation phase and initiates parallel verification of all previously generated draft tokens. Crucially, since these tokens have their hidden states cached up to Layer $d_{\max}$, they can be jointly processed through the remaining layers ($d_{\max}+1$ to $L$) in a single parallel pass. If all prior tokens are accepted, the system resumes decoding from the correct position — effectively using the verification step to also decode the difficult token without wasting prior computation.

\paragraph{Handling Long Shallow Chains}
Another case BSCS handles is when $w_{\max}$ consecutive tokens exit successfully at shallow layers, the risk of cumulative errors increases as each subsequent draft relies on the previous potentially unreliable predictions. To mitigate this, BSCS also triggers an early verification round. The operation is the same as that used for difficult tokens, except that it first aligns all $w_{\max}$ tokens to the deepest exit layer among them by running their missing layers in parallel which ensuring consistent context before batch verification. This mechanism controls error propagation, ensuring high accuracy for the next draft sequence. Due to space limitation, a detailed illustration of this case is provided in Appendix.

\paragraph{Hidden-State Cache Management}
To enable the draft interruption and verification triggering described above, BSCS maintains a hidden-state cache that stores intermediate representations. As depicted in Figure~\ref{dmax architecture}, the Cache Manager support two core operations: Write and Reuse. In the Write phase, when token $t_i$ exits at layer $\ell$, its hidden state $\mathbf{h}_i^{(\ell)}$ is stored in the cache. In the Reuse phase, cached hidden states are retrieved if they are needed, either when subsequent tokens require the missing KV cache of earlier tokens \cite{vaswani2017attention} or when batch verification begins, at which point the hidden states cached at layer $d_{\max}$ are concatenated and processed together through the deeper layers. This design guarantees that every output token undergoes computation across all layers, enabling lossless acceleration.

In summary, BSCS transforms speculative decoding from an unbounded, token-level process into a bounded, block-wise pipeline — balancing speed, reliability, and resource efficiency.

\subsection{Wall-time Speedup Analysis}
\label{subsec:Wall-time Speedup Analysis}
To demonstrate the effectiveness of SpecBound, we theoretically model its speedup over standard autoregressive decoding and analyze how key parameters influence the acceleration performance. 

We define speedup as the ratio of decoding throughput between SpecBound and autoregressive (AR) decoding, where throughput is measured as the number of tokens generated per unit wall-clock time. Let $T_{\text{AR}}$ denote the time required by AR decoding to generate one token (equivalent to a full forward pass through all $L$ layers). Our goal is to compare the expected token generation throughput of SpecBound against the AR baseline $1 / T_{\text{AR}}$.

\begin{table*}[t]
  \centering
  \small
  \begin{tabular}{lc@{\hspace{0.18cm}}c@{\hspace{0.2cm}}c@{\hspace{0.18cm}}c@{\hspace{0.2cm}}c@{\hspace{0.18cm}}c@{\hspace{0.2cm}}c@{\hspace{0.18cm}}c@{\hspace{0.2cm}}c@{\hspace{0.18cm}}c@{\hspace{0.2cm}}c@{\hspace{0.18cm}}c@{\hspace{0.2cm}}c}
    \toprule
    \multirow{2}[2]{*}{\textbf{Method}} &
      \multicolumn{2}{c@{\hspace{0.55cm}}}{\textbf{Math}} &
      \multicolumn{2}{c@{\hspace{0.55cm}}}{\textbf{Multi.}} &
      \multicolumn{2}{c@{\hspace{0.55cm}}}{\textbf{QA}} &
      \multicolumn{2}{c@{\hspace{0.55cm}}}{\textbf{RAG}} &
      \multicolumn{2}{c@{\hspace{0.55cm}}}{\textbf{Sum.}} &
      \multicolumn{2}{c@{\hspace{0.55cm}}}{\textbf{Trans.}} &
      \multirow{2}[2]{*}{\textbf{Overall}} \\
    \cmidrule(lr){2-13}
    & CR & SD & CR & SD & CR & SD & CR & SD & CR & SD & CR & SD & \\
    \midrule
    \multicolumn{14}{c}{\textbf{Vicuna-7B}} \\
    \midrule
    Lookahead \cite{fu2024break}  & 1.92 & 1.54$\times$ & 1.74 & 1.49$\times$ & 1.52 & 1.23$\times$ & 1.53 & 1.24$\times$ & 1.57 & 1.38$\times$ & 1.28 & 1.16$\times$ & 1.35$\times$ \\
    Medusa \cite{cai2023medusa}   & 1.77 & \textbf{1.96$\times$} & 1.72 & \textbf{1.95$\times$} & 1.54 & 1.67$\times$ & 1.47 & 1.47$\times$ & 1.53 & 1.59$\times$ & 1.55 & 1.63$\times$ & 1.71$\times$ \\
    REST \cite{he2024rest}        & 1.49 & 1.19$\times$ & 2.04 & 1.65$\times$ & 1.87 & 1.64$\times$ & 1.96 & 1.75$\times$ & 1.60 & 1.34$\times$ & 1.58 & 1.33$\times$ & 1.47$\times$ \\
    Kangaroo \cite{liu2024kangaroo}                  & 2.14 & 1.61$\times$ & 2.22 & 1.68$\times$ & 1.87 & 1.43$\times$ & 2.05 & 1.52$\times$ & 1.87 & 1.50$\times$ & 1.41 & 1.24$\times$ & 1.50$\times$ \\
    SPACE \cite{yi2024generation}                     & 2.15 & 1.65$\times$ & 2.31 & 1.71$\times$ & 1.72 & 1.33$\times$ & 1.88 & 1.24$\times$ & 2.19 & 1.47$\times$ & 1.73 & 1.37$\times$ & 1.46$\times$ \\
    Ours                      & \textbf{3.78} & 1.89$\times$ & \textbf{3.05} & 1.65$\times$ & \textbf{4.23} & \textbf{2.16$\times$} & \textbf{5.48} & \textbf{2.31$\times$} & \textbf{3.97} & \textbf{1.95$\times$} & \textbf{6.41} & \textbf{2.94$\times$} & \textbf{2.15$\times$} \\
    \midrule
    \multicolumn{14}{c}{\textbf{Vicuna-13B}}\\
    \midrule
    Lookahead \cite{fu2024break} & 1.98 & 1.61$\times$ & 1.64 & 1.41$\times$ & 1.43 & 1.20$\times$ & 1.48 & 1.19$\times$ & 1.54 & 1.30$\times$ & 1.22 & 1.07$\times$ & 1.30$\times$ \\
    Medusa \cite{cai2023medusa}  & 1.76 & \textbf{2.06$\times$} & 1.86 & \textbf{2.07$\times$} & 1.53 & 1.71$\times$ & 1.49 & 1.59$\times$ & 1.57 & 1.64$\times$ & 1.65 & 1.73$\times$ & 1.81$\times$ \\
    REST \cite{he2024rest}       & 1.59 & 1.21$\times$ & 1.94 & 1.50$\times$ & 1.96 & 1.55$\times$ & 1.83 & 1.53$\times$ & 1.62 & 1.35$\times$ & 1.57 & 1.19$\times$ & 1.37$\times$ \\
    Kangaroo \cite{liu2024kangaroo}                 & 2.42 & 1.63$\times$ & 2.44 & 1.66$\times$ & 1.79 & 1.34$\times$ & 2.16 & 1.40$\times$ & 2.00 & 1.41$\times$ & 1.45 & 1.18$\times$ & 1.44$\times$ \\
    Ours                     & \textbf{4.09} & 1.91$\times$ & \textbf{3.22} & 1.69$\times$ & \textbf{4.54} & \textbf{2.18$\times$} & \textbf{5.25} & \textbf{2.26$\times$} & \textbf{4.73} & \textbf{2.18$\times$} & \textbf{5.93} & \textbf{2.77$\times$} & \textbf{2.16$\times$} \\
    \midrule
    \multicolumn{14}{c}{\textbf{Codellama-7B-Instruct}} \\
    \midrule
    Lookahead \cite{fu2024break}              & 1.88 & 1.52$\times$ & 1.78 & 1.51$\times$ & 1.49 & 1.21$\times$ & 1.57 & 1.26$\times$ & 1.45 & 1.32$\times$ & 1.32 & 1.18$\times$ & 1.33$\times$ \\
    Medusa \cite{cai2023medusa}                 & 1.78 & 1.96$\times$ & 1.68 & \textbf{1.92$\times$} & 1.58 & 1.69$\times$ & 1.44 & 1.45$\times$ & 1.57 & 1.61$\times$ & 1.51 & 1.61$\times$ & 1.70$\times$ \\
    REST \cite{he2024rest}                   & 1.45 & 1.17$\times$ & 2.07 & 1.67$\times$ & 1.84 & 1.62$\times$ & 1.99 & 1.77$\times$ & 1.57 & 1.32$\times$ & 1.62 & 1.35$\times$ & 1.47$\times$ \\
    AdaDecode \cite{wei2025adadecode}               & 2.20 & 1.62$\times$ & 2.00 & 1.40$\times$ & 2.30 & 1.50$\times$ & 2.50 & 1.48$\times$ & 2.15 & 1.38$\times$ & 2.40 & 1.52$\times$ & 1.45$\times$ \\
    Ours                   & \textbf{3.63} & \textbf{2.02$\times$} & \textbf{2.74} & 1.66$\times$ & \textbf{3.13} & \textbf{1.84$\times$} & \textbf{3.98} & \textbf{1.88$\times$} & \textbf{2.98} & \textbf{1.78$\times$} & \textbf{4.74} & \textbf{2.42$\times$} & \textbf{1.93$\times`$} \\
    \midrule
    \multicolumn{14}{c}{\textbf{Codellama-13B-Instruct}} \\
    \midrule
    Lookahead \cite{fu2024break}             & 2.02 & 1.63$\times$ & 1.61 & 1.39$\times$ & 1.46 & 1.22$\times$ & 1.45 & 1.17$\times$ & 1.57 & 1.32$\times$ & 1.18 & 1.05$\times$ & 1.31$\times$ \\
    Medusa \cite{cai2023medusa}                & 1.74 & 2.04$\times$ & 1.89 & \textbf{2.09$\times$} & 1.42 & 1.65$\times$ & 1.60 & 1.65$\times$ & 1.47 & 1.58$\times$ & 1.68 & 1.75$\times$ & 1.81$\times$ \\
    REST \cite{he2024rest}                  & 1.63 & 1.23$\times$ & 1.91 & 1.48$\times$ & 1.99 & 1.57$\times$ & 1.80 & 1.51$\times$ & 1.73 & 1.41$\times$ & 1.46 & 1.13$\times$ & 1.39$\times$ \\
    AdaDecode \cite{wei2025adadecode}              & 2.40 & 1.75$\times$ & 2.20 & 1.52$\times$ & 2.50 & 1.60$\times$ & 2.70 & 1.58$\times$ & 2.35 & 1.48$\times$ & 2.60 & 1.65$\times$ & 1.52$\times$ \\
    Ours                  & \textbf{3.49} & \textbf{2.20$\times$} & \textbf{2.69} & 1.85$\times$ & \textbf{2.97} & \textbf{1.97$\times$} & \textbf{3.82} & \textbf{3.25$\times$} & \textbf{3.00} & \textbf{2.13$\times$} & \textbf{4.02} & \textbf{2.56$\times$} & \textbf{2.33$\times$} \\
    \bottomrule
  \end{tabular}
  \caption{Performance comparison of SpecBound against other self-drafting methods across multiple base models and diverse text generation tasks on Spec-Bench. We report compression rate (CR, average accepted tokens per draft round) and wall-time speedup (SD) relative to autoregressive decoding.}
  \label{tab:benchmark}
\end{table*}
\paragraph{Variables Definition}  
We model the decoding throughput of SpecBound using two parameters: $w$, the number of draft tokens generated per speculation round ($1 \leq w \leq w_{\max}$); and $\alpha$, the per-token acceptance rate—the probability that a draft token is accepted during verification. Note that $d_{\max}$, $w_{\max}$, and $L$ have been defined in earlier sections. 

\paragraph{Calculate Speedup Ratio}
In our method, a speculation round comprises two phases:
(1) Draft phase: sequential generation of $w$ tokens, each with $d_{\max}$ layers;  
(2) Verification phase: parallel processing of all $w$ tokens through the remaining $L - d_{\max}$ layers.  
Thus, total time per round is:
\begin{equation}
  \label{eq:wall-time analysis1}
    T_{\text{round}} = \frac{w d_{\max} + L - d_{\max}}{L} \cdot T_{\text{AR}}
\end{equation}

Not all draft tokens will be accepted every round. Under the standard geometric-acceptance assumption~\cite{leviathan2023fast}, the expected number of accepted tokens per round $\mathbb{E}[N_{\text{acc}}]$ is:  
\begin{equation}
\label{eq:wall-time analysis2}
   \sum_{k=0}^{w-1} k \alpha^k (1 - \alpha) + w \alpha^w = \frac{\alpha (1 - \alpha^w)}{1 - \alpha}
\end{equation}

Then our method's throughput is $\mathbb{E}[N_{\text{acc}}] / T_{\text{round}}$. Dividing by the AR decoding throughput $1 / T_{\text{AR}}$ and simplifying, we obtain the final speedup ratio (denoted as SD):
\begin{equation}
\label{eq:wall-time analysis3}
  \text{SD} = \frac{L \alpha (1 - \alpha^w)}{(1 - \alpha) \left( w d_{\max} + L - d_{\max} \right)}.
\end{equation}

\paragraph{Parameter Analysis}
From Equation~\eqref{eq:wall-time analysis3}, we observe that the overall speedup increases monotonically with the token acceptance rate $\alpha$. The Annealed Confidence Threshold early exit mechanism improves $\alpha$ by suppressing spurious tokens at shallow layers. Moreover, speedup can also be enhanced by increasing $w$ (or $w_{\max}$) or reducing $d_{\max}$, both of which lower the per-round time cost; however, such choices may compromise generation quality and, in turn, potentially reduce $\alpha$, reflecting an empirical trade-off that must be carefully balanced. Detailed experimental analysis of these hyperparameters is provided in Section~\ref{ablation study}.

%% file: tex/4-experiment.tex
\section{Experiments}
\subsection{Experimental Setup}
\label{Experimental Setup}
\paragraph{Backbone Models and Training}
We choose Vicuna~\cite{chiang2023vicuna,touvron2023llama} and CodeLlama-Instruct~\cite{roziere2023code} models with 7B and 13B parameters as backbone LLMs. During training, all base model parameters are frozen; only a lightweight LM head is trained per intermediate layer to enable early exit\cite{wei2025adadecode}. The LM heads are trained for 20 epochs on 68K multi-turn conversations from the ShareGPT dataset, using the AdamW optimizer with learning rate $3\times10^{-5}$ which taking approximately 2 hours on four NVIDIA H800 GPUs.

\paragraph{Evaluation Benchmarks and Baselines}
We evaluate our method on Spec-Bench, a diverse benchmark covering machine translation, question answering, and other text generation tasks\cite{xia2024unlocking}. We compare against a range of established self-drafting approaches, including Lookahead~\cite{fu2024break}, Medusa~\cite{cai2023medusa}, REST~\cite{he2024rest}, Kangaroo~\cite{liu2024kangaroo}, SPACE~\cite{yi2024generation}, and AdaDecode~\cite{wei2025adadecode}.Performance is measured using two key metrics: (1) compression rate (CR), defined as the average number of accepted tokens per speculation round, and (2) wall-time speedup (SD) relative to standard autoregressive decoding. For inference with our method, we set $w_{\max}=8$, $d_{\max}=10$, $\alpha=0.2$, and $\gamma=0.55$—a configuration that balances draft consumption and generation accuracy; a detailed exploration of inference parameters is provided in the ablation study. For fair comparison, all methods are evaluated on a single NVIDIA H800 GPU. All results are presented in Table~\ref{tab:benchmark}.
\subsection{Main Results}

\begin{figure*}[t]
  \includegraphics[width=1\linewidth]{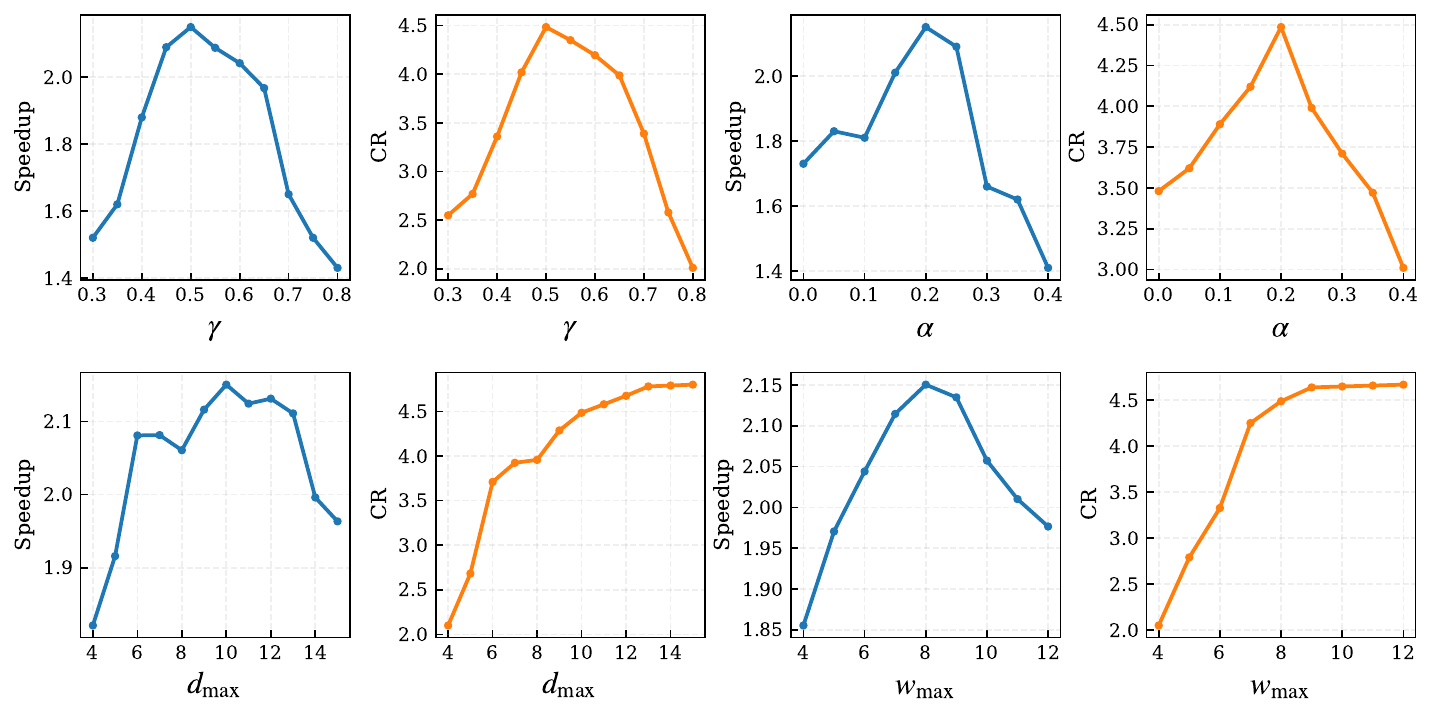}
    \caption{Hyperparameter Sensitivity Analysis on Vicuna-7B: wall-time speedup (blue) and compression rate (CR, orange) under varying exit threshold $\gamma$, annealing coefficient $\alpha$, depth bound $d_{\max}$, and width bound $w_{\max}$.}
  \label{ablation_para}
\end{figure*}
\begin{figure}[t]
  \includegraphics[width=1\linewidth]{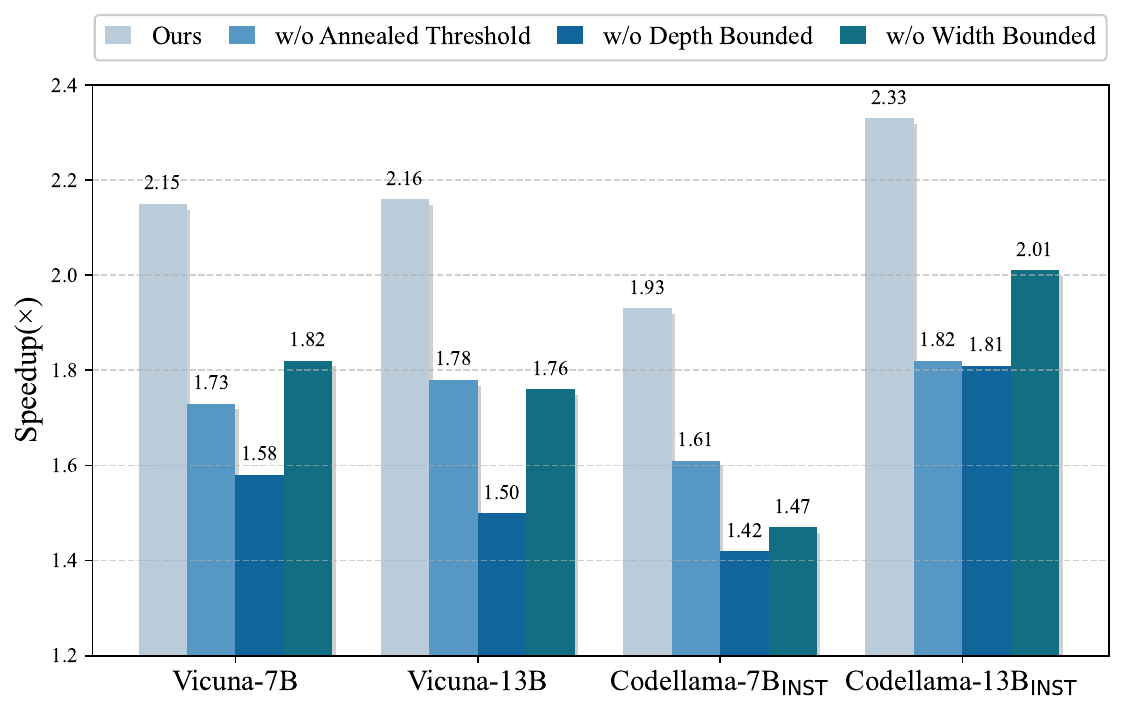}
    \caption{Ablation study of SpecBound components on wall-time speedup across different base models, comparing the full method against variants that removing the Annealed Confidence Threshold early exit mechanism, the depth bound $d_{\max}$, or the width bound $w_{\max}$.}
  % \caption {A minimal working example to demonstrate how to place
  %   two images side-by-side.}
  \label{ablation_arch}
\end{figure}

\paragraph{SpecBound achieves better overall speedup performance across tasks and base models.}
As shown in Table \ref{tab:benchmark}, our method achieves strong performance across diverse tasks, yielding the best overall speedup compared to existing approaches. Notably, it excels on tasks such as translation, where it attains speedup up to $2.94\times$, with consistently competitive results across some other tasks such as multi-turn dialogue. This performance variation likely stems from task-dependent early-exit behaviors: the inference hyperparameters selected to maximize overall speedup may not be optimal for every individual task. In practice, task-specific tuning of these parameters could further improve acceleration efficiency.

\paragraph{SpecBound enables better speedup through high-quality drafting.} Our method generates draft tokens using multiple transformer layers, incurring higher computational cost per draft step compared to lightweight alternatives (e.g., linear heads or single-layer transformers). However, this investment yields significantly more accurate drafts — as evidenced by our consistently highest acceptance rate (CR $\geq 3.0$ across all tasks), which enables deeper and more efficient multi-token speculation. Crucially, because each draft token is computed only up to a shallow or intermediate layer, the per-token draft cost remains substantially lower than that of a complete forward pass through the base LLM. As a result, the time saved from fewer speculative rejections more than compensates for the additional draft generation cost, resulting in superior overall speedup. 
This presents an alternative design principle: while most prior work focuses on minimizing draft latency, we prioritize draft quality to maximize end-to-end throughput.

\paragraph{Larger models benefit more from early exit despite similar draft quality.}
Interestingly, SpecBound consistently achieves higher speedup on 13B models than their 7B counterparts—even when compression rates (CR) are comparable. This suggests that deeper architectures offer greater acceleration potential through early exit, as a larger fraction of their deep layers are redundant for most easy tokens, enabling early-exited drafts to be verified in parallel over more layers. As a result, wall-time savings per accepted token are amplified, leading to higher net speedup even when CR plateaus.

\paragraph{Performance Robustness Across Model Domains}
% Furthermore, Vicuna—being a chat-optimized model—achieves consistently higher compression rates (CR) across all Spec-Bench tasks, which are predominantly natural-language oriented. In contrast, CodeLlama, fine-tuned primarily on code datasets, exhibits relatively lower CR on these general-language tasks, underscoring the impact of domain alignment between the base model and the evaluation benchmark. Nevertheless, SpecBound consistently delivers the highest speedup across both model families, demonstrating its robustness to architectural and domain differences.
Furthermore, as a chat-optimized model, Vicuna achieves consistently higher compression rates (CR) across all Spec-Bench tasks, which are predominantly natural-language oriented. In contrast, CodeLlama-Instruct, fine-tuned primarily on code datasets, exhibits relatively lower CR on these general-language tasks, underscoring the impact of domain alignment between the base model and the evaluation benchmark. Nevertheless, SpecBound consistently delivers the highest speedup across both model families, demonstrating its robustness to architectural and domain differences without any modification to the base LLM parameters. 
% Notably, this strong performance is achieved , highlighting the plug-and-play nature of our approach.
\begin{table*}[t]
  \centering
  % \small
  \begin{tabular}{l@{\hspace{1.0cm}}c@{\hspace{0.4cm}}c@{\hspace{0.4cm}}c@{\hspace{0.4cm}}c@{\hspace{0.4cm}}c@{\hspace{0.4cm}}c@{\hspace{0.4cm}}c}
    \toprule
    \textbf{Method} & \textbf{Math} & \textbf{Multi.} & \textbf{QA} & \textbf{RAG} & \textbf{Sum.} & \textbf{Trans.} & \textbf{Overall} \\
    \midrule
    greedy decoding & \textbf{1.89$\times$} & \textbf{1.65$\times$} & \textbf{2.16$\times$} & \textbf{2.31$\times$} & \textbf{1.95$\times$} & \textbf{2.94$\times$} & \textbf{2.15$\times$} \\
    temperature sampling (T=0.3) & 1.82$\times$ & 1.56$\times$ & 2.01$\times$ & 2.20$\times$ & 1.86$\times$ & 2.86$\times$ & 2.08$\times$ \\
    temperature sampling (T=1.0) & 1.71$\times$ & 1.44$\times$ & 1.92$\times$ & 2.08$\times$ & 1.73$\times$ & 2.63$\times$ & 1.88$\times$ \\
    top-p sampling (p=0.9) & 1.78$\times$ & 1.48$\times$ & 1.99$\times$ & 2.15$\times$ & 1.78$\times$ & 2.71$\times$ & 1.94$\times$ \\
    \bottomrule
  \end{tabular}
  \caption{Speedup performance of SpecBound (Vicuna-7B) under different decoding strategies on different tasks of Spec-Bench. We choose greedy docoding, temperature sampling and top-p sampling with different settings.}
  \label{tab:decoding}
\end{table*}
\subsection{Ablation Study}
\label{ablation study}
To demonstrate the robustness of our method and identify its effective hyperparameter range, we conduct a series of ablation studies on Vicuna-7B using the Spec-Bench benchmark. In each experiment, we vary one parameter—exit threshold $\gamma$, annealing coefficient $\alpha$, depth bound $d_{\max}$, or width bound $w_{\max}$—while fixing all others at their optimal values. The resulting wall-time speedup (blue) and compression rate (CR, orange) are shown in Figure~\ref{ablation_para}. We further conduct an ablation study to assess the contribution of each component in SpecBound. The experiment evaluates the full method against three variants—each removing one key component: the Annealed Confidence Threshold, the depth bound , or the width bound—across multiple base models on the Spec-Bench benchmark, as shown in Figure~\ref{ablation_arch}.
\paragraph{Impact of Annealed Confidence Threshold Parameters}
Figure~\ref{ablation_para} (top) shows that both the exit threshold $\gamma$ and annealing coefficient $\alpha$ critically affect performance. A low $\gamma$ may admit spurious shallow token exits, while a high $\gamma$ suppresses valid token exits. Similar trade-offs arise with the annealing strength: a small $\alpha$ fails to deal with spurious overconfidence, whereas a large $\alpha$ over-smooths the distribution, hindering early exit. Generally, the best speedup is achieved with a moderate $\gamma$ and a small $\alpha$, which mildly adjusts the softmax to balance reliability and efficiency.

\paragraph{Impact of Depth and Width Bounds}
Figure~\ref{ablation_para} (bottom) shows that speedup initially rises with $d_{\max}$ and $w_{\max}$, as more draft tokens can be generated and verified per speculation step. However, excessive $d_{\max}$ reduce the proportion of computation that can benefit from parallel shallow-depth processing. Similarly, an overly large $w_{\max}$ cause the errors to accumulate. These effects diminish end-to-end  throughput despite increased the length of accepted tokens. The optimal balance is achieved at $d_{\max}=10$ and $w_{\max}=8$, maximizing speculative efficiency without frequent interruption.

\paragraph{Component Contributions in Ablation Study}
Figure~\ref{ablation_arch} shows that removing any component of SpecBound leads to performance degradation, confirming their collective contribution to speedup. Removing the depth bound ($d_{\max}$) causes the largest drop, highlighting its critical role in mitigating difficult tokens by capping speculative depth and preserving parallelism across deeper layers. Removing the width bound ($w_{\max}$) results in a smaller decline: when removed, the system may still benefit from $d_{\max}$, as longer draft sequences tend to push exit layers deeper due to increasing model uncertainty, naturally triggering early verification. Nevertheless, $w_{\max}$ remains essential as the primary gate against excessively long shallow draft sequences, ensuring reliability without sacrificing efficiency.
\begin{table}[t]
  \centering
  \small
  \begin{tabular}{l@{\hspace{0.4cm}}c@{\hspace{0.4cm}}c@{\hspace{0.4cm}}c}
    \toprule
    \textbf{Metric} & \textbf{AR (Base)} & \textbf{Ours} & \textbf{Improv.} \\
    \midrule
    TTFT (ms) & 20.8 & 21.1 & - \\
    Per-token latency (ms) & 17.5 & 8.8 & \textbf{1.98$\times$} \\
    Throughput (tokens/s) & 57.3 & 123.1 & \textbf{2.15$\times$} \\
    End-to-End Latency (s) & 3.31 & 1.89 & \textbf{1.75$\times$} \\
    \bottomrule
  \end{tabular}
  \caption{Serving latency metrics comparison on Vicuna-7B using a single NVIDIA H800 GPU. "AR" denotes standard autoregressive decoding.}
  \label{tab:latency}
\end{table}
\paragraph{Effect of Sampling-Based Decoding Strategies}
To further evaluate the versatility of SpecBound, we conduct experiments on Spec-Bench using Vicuna-7B to assess its performance under various sampling-based decoding strategies, including temperature sampling ($T=0.3$ and $1.0$) and top-$p$ sampling ($p=0.9$). As summarized in Table~\ref{tab:decoding}, SpecBound consistently delivers significant wall-time speedup across all tested regimes, ranging from $1.88\times$ to $2.08\times$, even as sampling randomness increases. While the peak acceleration of $2.15\times$ is observed under greedy decoding, the performance remains robust across diverse stochastic settings. The slight moderation in speedup at higher temperatures is expected, as increased token entropy inherently poses challenges for speculative acceptance. These results demonstrate that SpecBound is not restricted to greedy search and can effectively accelerate various practical decoding configurations in real-world applications.

\paragraph{Detailed Serving Latency Analysis}
To provide a more actionable view of SpecBound's practical utility, we evaluate absolute serving latency metrics using Vicuna-7B on SpecBench with NVIDIA H800 GPUs, maintaining a fixed maximum length of 1024 and batch size of 1. As summarized in Table~\ref{tab:latency}, SpecBound achieves significant improvements in per-token streaming latency and throughput over the autoregressive baseline. Notably, the prefill phase remains identical to the base model, resulting in a Time to First Token (TTFT) nearly equal to the baseline. The marginal difference is attributed to minimal one-time operations such as variable initialization and the allocation of the small, fixed-size cache table ($L \times w_{\max}$). Furthermore, because the cache management involves minimal hardware operations and the table is reused for each draft window rather than growing with the input, SpecBound is not negatively impacted by long-context tasks in terms of memory overhead.

%% file: tex/2-background.tex
\section{Related Work}

Recent advancements in accelerating Large Language Models (LLMs) stem from the draft-then-verify paradigm, first introduced by Blockwise Decoding~\cite{stern2018blockwise}, which later inspired Speculative Decoding~\cite{leviathan2023fast} and Speculative Sampling~\cite{chen2023accelerating}. Building upon speculative decoding with an independent draft model, recent works explore self-draft strategies that leverage the base LLM itself for speculation. These Self-draft methods can be broadly categorized into three paradigms: (1) fine-tuning (2) layer skip and (3) early exit.

\subsection{Fine-tuning Methods}
Some works fine-tune the base LLM on non-autoregressive objectives to endow it with parallel token prediction capabilities~\cite{lin2025bita,yi2024generation}. However, such parameter updates not only require substantial computational resources for training but also risk degrading the model's original generalization performance on diverse downstream tasks. In contrast, our approach requires no fine-tuning and keeps all parameters frozen, ensuring that the model's inherent capabilities remain intact while significantly reducing the barrier to deployment. This plug-and-play nature allows SpecBound to be seamlessly integrated with existing pre-trained LLMs without any additional training overhead or performance trade-offs.

\subsection{Layer Skip Methods}
Another line of work accelerates drafting by using only a subset of the LLM’s layers—similar to layer pruning~\cite{kim2024shortened,zeng2023learning,yang2024laco,men2025shortgpt}. For example, \cite{zhang2024draft} adaptively skips intermediate layers during draft generation. However, since skipped layers are never executed for draft tokens, their hidden states remain incomplete, complicating rigorous verification and risking output correctness. Furthermore, such skipping often leads to a mismatch in the KV cache across layers. In contrast, our method ensures every output token passes through all base LLM layers during verification, preserving distributional fidelity. By maintaining a continuous and complete computation flow, SpecBound eliminates the risks associated with missing intermediate representations, thereby guaranteeing that the accelerated output is mathematically identical to that of the original model.

\subsection{Early Exit Methods}
To preserve per-layer computation, early exit methods adopt the strategy of producing draft tokens by executing only the initial few layers of the base LLM, and then feeding these drafts into the subsequent decoding step for full-depth generation \cite{bolukbasi2017adaptive,li2021accelerating,kong2022accelerating,varshney2023accelerating}.
Notably, AdaDecode~\cite{wei2025adadecode} also adopts an early-exit plus verification strategy similar in spirit to ours. However, its draft generation relies on a simplistic early-exit criterion and imposes no bound on speculation depth, which limits its acceleration potential. In contrast, SpecBound introduces a more refined early-exit mechanism to improve draft accuracy and Bounded Speculation with Cached States Algorithm to maximize parallel efficiency—enabling higher speedup while preserving exact output equivalence.

%% file: tex/5-conclusion.tex
\section{Conclusion}
We present SpecBound, a self-draft speculative decoding method that explicitly addresses the heterogeneity in early-exit difficulty across tokens during decoding. To mitigate spurious high-confidence predictions in shallow layers, we introduce a carefully designed annealed temperature-sampling early exit judge mechanism. To prevent redundant computation on difficult tokens and error propagation in long shallow draft chains, we propose a cache-based draft-verification algorithm with adaptive depth and width bounds. Notably, SpecBound achieves strong and consistent speedup across diverse base models and text generation tasks, while ensuring lossless acceleration without any modification to the base LLM parameters.

%% file: tex/6-limitation.tex
\section*{Limitations}
Our approach, while avoiding the costly training of a separate draft model, still requires lightweight trainable heads at intermediate layers to enable accurate early exiting, and is therefore not fully training-free. Future work could explore mechanisms for reliable shallow exits without any parameter updates. Moreover, the method involves several hyperparameters whose optimal settings currently rely on empirical tuning. A promising direction is to develop adaptive or tuning-free strategies, e.g., dynamically adjusting speculation bounds based on online token-level uncertainty, to enhance robustness and ease of deployment.

%% file: tex/8-Acknowledgements.tex
\section*{Acknowledgements}
We gratefully acknowledge all the reviewers for their valuable comments and suggestions. This work
was supported by the Natural Science Foundation of Beijing, China (Grant No. L257006).

%% file: tex/7-appendix.tex
% \begin{figure*}[t]
%   % \includegraphics[width=0.48\linewidth]{example-image-a} \hfill
%   \includegraphics[width=1\linewidth]{ARR结构图_appendix.pdf}
%   \caption{\textbf{Left:} Overview of the BSCS algorithm under the setting $d_{\max}=6$ and $w_{\max}=4$. When token ``chat'' successfully exits at layer 2 and the total draft length reaches the width bound $w_{\max}$, speculation is terminated. All hidden states of previously generated tokens (``I'', ``am'', ``a'', ``chat'') are concatenated and passed to the remaining layers for parallel verification. \textbf{Right:} Detailed illustration of per-token layer-wise computation and hidden-state cache management.}
%     \label{wmax architecture}
% \end{figure*}
\section{Illustration of Long Shallow Chains}
Given a decoding history prefix ``Who are you?'', the base model early exits token ``I'' at layer 2, token ``am'' at layer 4, token ``a'' at layer 6, and token ``chat'' at layer 2 in four consecutive steps. After the fourth token, the draft length reaches the width bound $w_{\max}=4$, triggering an early verification despite all tokens having successfully exited at shallow layers. At this point, BSCS terminates the speculation phase and initiates parallel verification of all draft tokens. Crucially, because these tokens exited at different shallow layers, their hidden states are first aligned to the deepest exit layer among them (layer 6 in this case) by running the missing intermediate layers in parallel. The aligned representations are then jointly processed through the remaining layers (layer 7 to $L$) in a single forward pass, ensuring consistent context for batched verification. This alignment step prevents error propagation from shallow, potentially unreliable predictions while preserving the efficiency of multi-token speculation.

\section{Baseline Implementation Details}
For a fair comparison across all base models and tasks in Spec-Bench, we retrain the draft components of AdaDecode~\cite{wei2025adadecode} and Kangaroo~\cite{liu2024kangaroo} on the same 68K-shareGPT dataset used for SpecBound. Although public weights for these methods are available for certain model configurations (e.g., Vicuna-7B), they do not cover the full range of our evaluation settings(base models or benchmarks). All other baselines (Lookahead~\cite{fu2024break}, Medusa~\cite{cai2023medusa}, REST~\cite{he2024rest}, SPACE~\cite{yi2024generation}) are evaluated using their officially released weights and the standard Spec-Bench\cite{xia2024unlocking} pipeline without modification. All results are measured on a single NVIDIA H800 GPU, with speedup reported relative to autoregressive decoding using the same base LLM.

\section{Training Overhead Details}
\label{sec:training_details}
As detailed in Section~\ref{Experimental Setup}, the training process is highly efficient, requiring only 20 epochs on 68K multi-turn conversations from the ShareGPT dataset. This setup is completed in approximately two hours using four NVIDIA H800 GPUs.

A key factor in this efficiency is the "feature-caching" strategy employed during data generation. By storing the intermediate hidden states of the base LLM as it decodes the training inputs, the base model does not need to be re-run during the iterative training of the LM heads. Instead, we perform parallel supervision only on a set of lightweight linear layers, significantly reducing the computational burden. Once trained, these task-agnostic weights provide a permanent reduction in inference-time latency and resource consumption across various downstream applications.

\begin{table*}[t]
  \centering
  \small
  \begin{tabular}{l@{\hspace{1.5cm}}c@{\hspace{0.6cm}}c@{\hspace{0.6cm}}c@{\hspace{0.6cm}}c@{\hspace{0.6cm}}c@{\hspace{0.6cm}}c@{\hspace{0.6cm}}c}
    \toprule
    \textbf{Device} & \textbf{Math} & \textbf{Multi-turn} & \textbf{QA} & \textbf{RAG} & \textbf{Summary} & \textbf{Translation} & \textbf{Overall} \\
    \midrule
    H800 & 1.89$\times$ & 1.65$\times$ & 2.16$\times$ & 2.31$\times$ & 1.95$\times$ & 2.94$\times$ & \textbf{2.15$\times$} \\
    RTX 3090 & 1.81$\times$ & 1.58$\times$ & 2.12$\times$ & 2.13$\times$ & 1.92$\times$ & 2.69$\times$ & \textbf{2.01$\times$} \\
    \bottomrule
  \end{tabular}
  \caption{Speedup performance of SpecBound (Vicuna-7B) on different GPU devices across various Spec-Bench tasks. The evaluation are conducted on H800 and RTX 3090 with all other settings are the same.}
  \label{tab:hardware}
\end{table*}

\section{Empirical Robustness Across Different Hardware Architectures}
To validate the robustness of SpecBound across diverse hardware, we additionally evaluated our method on consumer-grade NVIDIA GeForce RTX 3090 GPUs (24GB), using Vicuna-7B as the base model. The RTX 3090 was selected for its more constrained memory subsystem—specifically its lower memory bandwidth and higher latency compared to the H800—to verify SpecBound's effectiveness under stricter hardware limitations. As shown in table \ref{tab:hardware}, while the overall speedup slightly decreases from $2.15\times$ to $2.01\times$ when switching to the RTX 3090, SpecBound remains consistently effective across all tasks. These results demonstrate the method's practical robustness and its suitability for diverse deployment scenarios, even on hardware with significant resource constraints.

\section{Robustness and Break-even Analysis}
\label{sec:robustness}

To address the potential concerns regarding the robustness in low-acceptance regimes (e.g., domain shift or highly ambiguous sequences), we provide a comprehensive analysis of its performance lower bound and empirical speedup distribution.

\paragraph{Theoretical Worst-Case Analysis}
Theoretically, the worst-case scenario for SpecBound occurs when the model speculates the maximum width $w_{\max}$ at depth $d_{\max}$, but all draft tokens are rejected during verification. In this case, only one token is successfully decoded in the cycle. Based on the speedup model in Eq. \ref{eq:wall-time analysis3}, as the per-token acceptance rate $a \to 0$ and speculation width $w \to w_{\max}$, the speedup ratio $SD$ can theoretically fall below $1.0$. However, SpecBound mitigates this "wasted" computation by reusing cached hidden states $h^{(d_{\max})}$ for the subsequent verification pass. Since the cost of a few early-layer passes is fractional compared to a full-model forward pass, the efficiency gap between consumption and gain is significantly narrowed, ensuring that the "investment" in drafting usually yields a net positive return.
\begin{table}[ht]
  \centering
  \begin{tabular}{l@{\hspace{0.3cm}}c@{\hspace{0.3cm}}c}
    \toprule
    \textbf{Speedup Range} & \textbf{Number} & \textbf{Percentage} \\
    \midrule
    $3.54\times - 4.16\times$ (Max) & 16 & 3.33\% \\
    $2.93\times - 3.54\times$ & 48 & 10.00\% \\
    $2.31\times - 2.93\times$ & 103 & 21.46\% \\
    $1.69\times - 2.31\times$ & 197 & 41.04\% \\
    $1.07\times$ (Min) $- 1.69\times$ & 116 & 24.17\% \\
    \bottomrule
  \end{tabular}
  \caption{Empirical speedup distribution of SpecBound on Vicuna-7B across 480 prompts from SpecBench.}
  \label{tab:distribution}
\end{table}
\paragraph{Empirical Distribution across Prompts}
To empirically validate the robustness, we analyze the per-prompt speedup distribution for Vicuna-7B on SpecBench (480 prompts). As summarized in Table~\ref{tab:distribution}, SpecBound consistently maintains a speedup above $1.0\times$ even in its tail cases. Specifically, 24.17\% of prompts fall into the minimum speedup range ($1.07\times - 1.69\times$), while over 34\% of prompts achieve a speedup exceeding $2.93\times$. The fact that the entire distribution remains above the break-even point ($SD > 1.0$) underscores the practical reliability of our adaptive bounded speculation, even when encountering difficult tokens or varying context lengths.

%% file: custom.bib
@article{wei2022chain,
  title={Chain-of-thought prompting elicits reasoning in large language models},
  author={Wei, Jason and Wang, Xuezhi and Schuurmans, Dale and Bosma, Maarten and Xia, Fei and Chi, Ed and Le, Quoc V and Zhou, Denny and others},
  journal={Advances in neural information processing systems},
  volume={35},
  pages={24824--24837},
  year={2022}
}

@article{achiam2023gpt,
  title={Gpt-4 technical report},
  author={Achiam, Josh and Adler, Steven and Agarwal, Sandhini and Ahmad, Lama and Akkaya, Ilge and Aleman, Florencia Leoni and Almeida, Diogo and Altenschmidt, Janko and Altman, Sam and Anadkat, Shyamal and others},
  journal={arXiv preprint arXiv:2303.08774},
  year={2023}
}

@article{touvron2023llama,
  title={Llama: Open and efficient foundation language models},
  author={Touvron, Hugo and Lavril, Thibaut and Izacard, Gautier and Martinet, Xavier and Lachaux, Marie-Anne and Lacroix, Timoth{\'e}e and Rozi{\`e}re, Baptiste and Goyal, Naman and Hambro, Eric and Azhar, Faisal and others},
  journal={arXiv preprint arXiv:2302.13971},
  year={2023}
}

@article{guo2025deepseek,
  title={Deepseek-r1: Incentivizing reasoning capability in llms via reinforcement learning},
  author={Guo, Daya and Yang, Dejian and Zhang, Haowei and Song, Junxiao and Zhang, Ruoyu and Xu, Runxin and Zhu, Qihao and Ma, Shirong and Wang, Peiyi and Bi, Xiao and others},
  journal={arXiv preprint arXiv:2501.12948},
  year={2025}
}

@article{shen2023hugginggpt,
  title={Hugginggpt: Solving ai tasks with chatgpt and its friends in hugging face},
  author={Shen, Yongliang and Song, Kaitao and Tan, Xu and Li, Dongsheng and Lu, Weiming and Zhuang, Yueting},
  journal={Advances in Neural Information Processing Systems},
  volume={36},
  pages={38154--38180},
  year={2023}
}

@article{chiang2023vicuna,
  title={Vicuna: An open-source chatbot impressing gpt-4 with 90\%* chatgpt quality},
  author={Chiang, Wei-Lin and Li, Zhuohan and Lin, Zi and Sheng, Ying and Wu, Zhanghao and Zhang, Hao and Zheng, Lianmin and Zhuang, Siyuan and Zhuang, Yonghao and Gonzalez, Joseph E and others},
  journal={See https://vicuna. lmsys. org (accessed 14 April 2023)},
  volume={2},
  number={3},
  pages={6},
  year={2023}
}

@inproceedings{leviathan2023fast,
  title={Fast inference from transformers via speculative decoding},
  author={Leviathan, Yaniv and Kalman, Matan and Matias, Yossi},
  booktitle={International Conference on Machine Learning},
  pages={19274--19286},
  year={2023},
  organization={PMLR}
}

@article{chen2023accelerating,
  title={Accelerating large language model decoding with speculative sampling},
  author={Chen, Charlie and Borgeaud, Sebastian and Irving, Geoffrey and Lespiau, Jean-Baptiste and Sifre, Laurent and Jumper, John},
  journal={arXiv preprint arXiv:2302.01318},
  year={2023}
}

@article{stern2018blockwise,
  title={Blockwise parallel decoding for deep autoregressive models},
  author={Stern, Mitchell and Shazeer, Noam and Uszkoreit, Jakob},
  journal={Advances in Neural Information Processing Systems},
  volume={31},
  year={2018}
}

@article{xia2024unlocking,
  title={Unlocking efficiency in large language model inference: A comprehensive survey of speculative decoding},
  author={Xia, Heming and Yang, Zhe and Dong, Qingxiu and Wang, Peiyi and Li, Yongqi and Ge, Tao and Liu, Tianyu and Li, Wenjie and Sui, Zhifang},
  journal={arXiv preprint arXiv:2401.07851},
  year={2024}
}

@inproceedings{xia2023speculative,
  title={Speculative decoding: Exploiting speculative execution for accelerating seq2seq generation},
  author={Xia, Heming and Ge, Tao and Wang, Peiyi and Chen, Si-Qing and Wei, Furu and Sui, Zhifang},
  booktitle={Findings of the Association for Computational Linguistics: EMNLP 2023},
  pages={3909--3925},
  year={2023}
}

@misc{cai2023medusa,
  title={Medusa: Simple framework for accelerating llm generation with multiple decoding heads},
  author={Cai, Tianle and Li, Yuhong and Geng, Zhengyang and Peng, Hongwu and Dao, Tri},
  year={2023}
}

@article{li2024eagle,
  title={Eagle: Speculative sampling requires rethinking feature uncertainty},
  author={Li, Yuhui and Wei, Fangyun and Zhang, Chao and Zhang, Hongyang},
  journal={arXiv preprint arXiv:2401.15077},
  year={2024}
}

@article{lin2025bita,
  title={BiTA: Bi-directional tuning for lossless acceleration in large language models},
  author={Lin, Feng and Yi, Hanling and Yang, Yifan and Li, Hongbin and Yu, Xiaotian and Lu, Guangming and Xiao, Rong},
  journal={Expert Systems with Applications},
  volume={279},
  pages={127305},
  year={2025},
  publisher={Elsevier}
}

@inproceedings{yi2024generation,
  title={Generation meets verification: Accelerating large language model inference with smart parallel auto-correct decoding},
  author={Yi, Hanling and Lin, Feng and Li, Hongbin and Peiyang, Ning and Yu, Xiaotian and Xiao, Rong},
  booktitle={Findings of the Association for Computational Linguistics: ACL 2024},
  pages={5285--5299},
  year={2024}
}

@inproceedings{zhang2024draft,
  title={Draft\& verify: Lossless large language model acceleration via self-speculative decoding},
  author={Zhang, Jun and Wang, Jue and Li, Huan and Shou, Lidan and Chen, Ke and Chen, Gang and Mehrotra, Sharad},
  booktitle={Proceedings of the 62nd Annual Meeting of the Association for Computational Linguistics (Volume 1: Long Papers)},
  pages={11263--11282},
  year={2024}
}

@inproceedings{elhoushi2024layerskip,
  title={Layerskip: Enabling early exit inference and self-speculative decoding},
  author={Elhoushi, Mostafa and Shrivastava, Akshat and Liskovich, Diana and Hosmer, Basil and Wasti, Bram and Lai, Liangzhen and Mahmoud, Anas and Acun, Bilge and Agarwal, Saurabh and Roman, Ahmed and others},
  booktitle={Proceedings of the 62nd Annual Meeting of the Association for Computational Linguistics (Volume 1: Long Papers)},
  pages={12622--12642},
  year={2024}
}

@article{wei2025adadecode,
  title={AdaDecode: Accelerating LLM Decoding with Adaptive Layer Parallelism},
  author={Wei, Zhepei and Chen, Wei-Lin and Zhu, Xinyu and Meng, Yu},
  journal={arXiv preprint arXiv:2506.03700},
  year={2025}
}

@article{liu2024kangaroo,
  title={Kangaroo: Lossless self-speculative decoding via double early exiting},
  author={Liu, Fangcheng and Tang, Yehui and Liu, Zhenhua and Ni, Yunsheng and Han, Kai and Wang, Yunhe},
  journal={arXiv preprint arXiv:2404.18911},
  year={2024}
}

@article{roziere2023code,
  title={Code llama: Open foundation models for code},
  author={Roziere, Baptiste and Gehring, Jonas and Gloeckle, Fabian and Sootla, Sten and Gat, Itai and Tan, Xiaoqing Ellen and Adi, Yossi and Liu, Jingyu and Sauvestre, Romain and Remez, Tal and others},
  journal={arXiv preprint arXiv:2308.12950},
  year={2023}
}

@article{fu2024break,
  title={Break the sequential dependency of llm inference using lookahead decoding},
  author={Fu, Yichao and Bailis, Peter and Stoica, Ion and Zhang, Hao},
  journal={arXiv preprint arXiv:2402.02057},
  year={2024}
}

@inproceedings{he2024rest,
  title={Rest: Retrieval-based speculative decoding},
  author={He, Zhenyu and Zhong, Zexuan and Cai, Tianle and Lee, Jason and He, Di},
  booktitle={Proceedings of the 2024 conference of the North American chapter of the association for computational linguistics: Human language technologies (volume 1: long papers)},
  pages={1582--1595},
  year={2024}
}

@article{gloeckle2024better,
  title={Better \& faster large language models via multi-token prediction},
  author={Gloeckle, Fabian and Idrissi, Badr Youbi and Rozi{\`e}re, Baptiste and Lopez-Paz, David and Synnaeve, Gabriel},
  journal={arXiv preprint arXiv:2404.19737},
  year={2024}
}

@article{chuang2023dola,
  title={Dola: Decoding by contrasting layers improves factuality in large language models},
  author={Chuang, Yung-Sung and Xie, Yujia and Luo, Hongyin and Kim, Yoon and Glass, James and He, Pengcheng},
  journal={arXiv preprint arXiv:2309.03883},
  year={2023}
}

@article{zhu2025layercake,
  title={LayerCake: Token-Aware Contrastive Decoding within Large Language Model Layers},
  author={Zhu, Jingze and Wu, Yongliang and Zhu, Wenbo and Cao, Jiawang and Zheng, Yanqiang and Chen, Jiawei and Yang, Xu and Schiele, Bernt and Fischer, Jonas and Hu, Xinting},
  journal={arXiv preprint arXiv:2507.04404},
  year={2025}
}

@inproceedings{geng2021learning,
  title={Learning to rewrite for non-autoregressive neural machine translation},
  author={Geng, Xinwei and Feng, Xiaocheng and Qin, Bing},
  booktitle={Proceedings of the 2021 Conference on Empirical Methods in Natural Language Processing},
  pages={3297--3308},
  year={2021}
}

@article{nie2025large,
  title={Large language diffusion models},
  author={Nie, Shen and Zhu, Fengqi and You, Zebin and Zhang, Xiaolu and Ou, Jingyang and Hu, Jun and Zhou, Jun and Lin, Yankai and Wen, Ji-Rong and Li, Chongxuan},
  journal={arXiv preprint arXiv:2502.09992},
  year={2025}
}

@article{zhang2025survey,
  title={A survey on parallel text generation: From parallel decoding to diffusion language models},
  author={Zhang, Lingzhe and Fang, Liancheng and Duan, Chiming and He, Minghua and Pan, Leyi and Xiao, Pei and Huang, Shiyu and Zhai, Yunpeng and Hu, Xuming and Yu, Philip S and others},
  journal={arXiv preprint arXiv:2508.08712},
  year={2025}
}

@inproceedings{bolukbasi2017adaptive,
  title={Adaptive neural networks for efficient inference},
  author={Bolukbasi, Tolga and Wang, Joseph and Dekel, Ofer and Saligrama, Venkatesh},
  booktitle={International conference on machine learning},
  pages={527--536},
  year={2017},
  organization={PMLR}
}

@inproceedings{li2021accelerating,
  title={Accelerating bert inference for sequence labeling via early-exit},
  author={Li, Xiaonan and Shao, Yunfan and Sun, Tianxiang and Yan, Hang and Qiu, Xipeng and Huang, Xuan-Jing},
  booktitle={Proceedings of the 59th annual meeting of the Association for Computational Linguistics and the 11th international joint conference on natural language processing (volume 1: long papers)},
  pages={189--199},
  year={2021}
}

@inproceedings{kong2022accelerating,
  title={Accelerating inference for pretrained language models by unified multi-perspective early exiting},
  author={Kong, Jun and Wang, Jin and Yu, Liang-Chih and Zhang, Xuejie},
  booktitle={Proceedings of the 29th International Conference on Computational Linguistics},
  pages={4677--4686},
  year={2022}
}

@article{varshney2023accelerating,
  title={Accelerating llm inference by enabling intermediate layer decoding},
  author={Varshney, Neeraj and Chatterjee, Agneet and Parmar, Mihir and Baral, Chitta},
  journal={CoRR},
  year={2023}
}

@article{kim2024shortened,
  title={Shortened llama: A simple depth pruning for large language models},
  author={Kim, Bo-Kyeong and Kim, Geonmin and Kim, Tae-Ho and Castells, Thibault and Choi, Shinkook and Shin, Junho and Song, Hyoung-Kyu},
  journal={arXiv preprint arXiv:2402.02834},
  volume={11},
  pages={1},
  year={2024}
}

@article{zeng2023learning,
  title={Learning to skip for language modeling},
  author={Zeng, Dewen and Du, Nan and Wang, Tao and Xu, Yuanzhong and Lei, Tao and Chen, Zhifeng and Cui, Claire},
  journal={arXiv preprint arXiv:2311.15436},
  year={2023}
}

@article{yang2024laco,
  title={Laco: Large language model pruning via layer collapse},
  author={Yang, Yifei and Cao, Zouying and Zhao, Hai},
  journal={arXiv preprint arXiv:2402.11187},
  year={2024}
}

@inproceedings{men2025shortgpt,
  title={Shortgpt: Layers in large language models are more redundant than you expect},
  author={Men, Xin and Xu, Mingyu and Zhang, Qingyu and Yuan, Qianhao and Wang, Bingning and Lin, Hongyu and Lu, Yaojie and Han, Xianpei and Chen, Weipeng},
  booktitle={Findings of the Association for Computational Linguistics: ACL 2025},
  pages={20192--20204},
  year={2025}
}

@article{vaswani2017attention,
  title={Attention is all you need},
  author={Vaswani, Ashish and Shazeer, Noam and Parmar, Niki and Uszkoreit, Jakob and Jones, Llion and Gomez, Aidan N and Kaiser, {\L}ukasz and Polosukhin, Illia},
  journal={Advances in neural information processing systems},
  volume={30},
  year={2017}
}

@article{xia2025tokenskip,
  title={Tokenskip: Controllable chain-of-thought compression in llms},
  author={Xia, Heming and Leong, Chak Tou and Wang, Wenjie and Li, Yongqi and Li, Wenjie},
  journal={arXiv preprint arXiv:2502.12067},
  year={2025}
}

@article{wen2024speculative,
  title={Speculative decoding with CTC-based draft model for LLM inference acceleration},
  author={Wen, Zhuofan and Gui, Shangtong and Feng, Yang},
  journal={Advances in Neural Information Processing Systems},
  volume={37},
  pages={92082--92100},
  year={2024}
}

@article{bu2026language,
  title={Language on Demand, Knowledge at Core: Composing LLMs with Encoder-Decoder Translation Models for Extensible Multilinguality},
  author={Bu, Mengyu and Feng, Yang},
  journal={arXiv preprint arXiv:2603.17512},
  year={2026}
}
